\documentclass[conference]{IEEEtran}
\IEEEoverridecommandlockouts

\usepackage{cite}
\usepackage[T1]{fontenc}
\usepackage[utf8]{inputenc}
\usepackage{microtype}
\usepackage{amsmath,amssymb,amsfonts}
\usepackage{mathtools}
\usepackage{graphicx}
\graphicspath{{figures/}}
\usepackage{booktabs}
\usepackage{array}
\usepackage{textcomp}
\usepackage{xcolor}
\usepackage{url}
\usepackage[hidelinks]{hyperref}

\newcommand{\HABL}{H_{\text{ABL}}}
\newcommand{\zmin}{z_{\min}}

\begin{document}

\title{Reconstructing Boundary Layer Wind Profiles with Machine Learning:\\
A Comparative Study on IGRA Radiosonde Data\thanks{Author manuscript of a paper accepted at the 2nd Workshop on AI-Ready Data for Science Discovery (ADSD), IEEE ICDM 2026. \copyright\ 2026 IEEE. Personal use of this material is permitted. Permission from IEEE must be obtained for all other uses, in any current or future media, including reprinting/republishing this material for advertising or promotional purposes, creating new collective works, for resale or redistribution to servers or lists, or reuse of any copyrighted component of this work in other works.}}


\author{%
\begin{tabular}{@{}c@{}}
\begin{tabular}{@{}c@{\hspace{2.5em}}c@{\hspace{2.5em}}c@{\hspace{2.5em}}c@{}}
\begin{tabular}[t]{@{}c@{}}
Fedor Buzaev\\
\textit{HSE University}\\
Moscow, Russia\\
fa.buzaev@hse.ru
\end{tabular} &
\begin{tabular}[t]{@{}c@{}}
Dmitry Efremenko\\
\textit{HSE University}\\
Moscow, Russia
\end{tabular} &
\begin{tabular}[t]{@{}c@{}}
Daria Pugacheva\\
Generalizable AI Systems\\
\textit{HSE University}\\
Moscow, Russia\\
dpugacheva@hse.ru
\end{tabular} &
\begin{tabular}[t]{@{}c@{}}
Fedor Ratnikov\\
\textit{HSE University}\\
Moscow, Russia
\end{tabular}
\end{tabular}\\[8.5ex]

\begin{tabular}{@{}c@{\hspace{2.5em}}c@{\hspace{2.5em}}c@{}}
\begin{tabular}[t]{@{}c@{}}
Egor Bugaev\\
\textit{HSE University}\\
Moscow, Russia
\end{tabular} &
\begin{tabular}[t]{@{}c@{}}
Vladimir Gordin\\
\textit{HSE University}\\
\textit{MIRAI}\\
Moscow, Russia
\end{tabular} &
\begin{tabular}[t]{@{}c@{}}
Denis Derkach\\
\textit{HSE University}\\
Moscow, Russia
\end{tabular}
\end{tabular}
\end{tabular}%
}
\maketitle

\begin{abstract}
Wind profiles in the atmospheric boundary layer (ABL) drive short-range
forecasting and wind-energy siting, yet sparse radiosonde coverage forces
practitioners to interpolate them from a handful of physical predictors.
We benchmark modern tabular learners against analytical parameterizations
on $158{,}097$ IGRA soundings. CatBoost, TabM, and FT-Transformer reduce
RMSE to $1.3$--$1.6$~m/s, against $\sim$3~m/s for power-law and
Monin--Obukhov fits; once the predictor set is fixed the three
architectures differ by less than $0.05$~m/s, so accuracy is driven by
predictors rather than architecture. A cohort analysis of within-bin
profile dispersion places an empirical floor of $\approx 1.77$~m/s on any
point-wise predictor that uses standard radiosonde features, consistent
with prior eddy-viscosity inversions. A one-dimensional Fourier Neural
Operator applied as a post-hoc corrector to frozen tabular predictions
reduces near-surface RMSE on U.S.\ stations from $1.5$ to $1.2$~m/s,
while leaving upper-layer error unchanged. Further gains should come
from new predictors rather than larger models.
\end{abstract}


\section{Introduction}

The atmospheric boundary layer (ABL) is the lowest part of the troposphere, where turbulent exchange of momentum, heat, and moisture governs surface--atmosphere interactions~\cite{stull1988boundary}. Accurate wind-profile reconstruction in the ABL is essential for short-range weather forecasting, pollutant transport, and wind-energy applications. Radiosondes provide high-resolution vertical soundings, yet their spatial and temporal coverage is sparse, which forces predictive models that infer complete wind profiles from a compact set of predictors.

Modeling wind in the ABL differs fundamentally from modeling in the free troposphere. Above the boundary layer, wind profiles are largely governed by large-scale geostrophic balance between Coriolis and pressure-gradient forces, which can be captured with relatively simple dynamical models. Within the ABL, turbulent mixing, surface friction, and stratification strongly modify the vertical wind structure. These processes introduce nonlinear and regime-dependent behavior, making boundary-layer profiles far more difficult to approximate with analytical parameterizations. Predictors that are sufficient for upper-air modeling are inadequate near the surface, where local conditions, diurnal cycles, and surface heterogeneity dominate. This complexity motivates flexible data-driven methods that can capture fine-scale dependencies while remaining grounded in physical predictors such as geostrophic wind, latitude, and boundary-layer depth.

Classical analytical approaches---the power-law profile and Monin--Obukhov similarity theory~\cite{monin1954basic}---capture first-order effects of surface roughness and stability but consistently underperform compared to radiosonde observations. Modern PBL parameterizations used in numerical weather prediction, such as the Yonsei University scheme~\cite{hong2006vertical}, refine these approximations but still rely on simplified closures that cannot fully represent the diversity of boundary-layer regimes observed in practice.

Hybrid data-driven approaches in Earth system science~\cite{reichstein2019deep} now train end-to-end on radiosonde and reanalysis data, bypassing fixed-form closures. Transformer-based models~\cite{attention}, modern tabular architectures such as TabM~\cite{gorishniy2025tabm}, and gradient boosting methods such as CatBoost~\cite{prokhorenkova2018catboost} have shown strong performance on structured prediction, which makes them the natural starting point for our benchmark.

\paragraph{Contributions.}
We make three contributions.
\begin{itemize}
  \item \textbf{A predictor-vs-architecture decomposition on a global IGRA benchmark.} On $158{,}097$ soundings we evaluate FT-Transformer, CatBoost, and TabM under a canonical and an extended predictor set~\cite{durre2006igra,durre2018enhancing}. Switching the predictor set moves RMSE from $\sim$2.5~m/s to $\sim$1.3--1.6~m/s; switching the architecture moves it by less than $0.05$~m/s. A leave-one-block-out ablation localizes the gain to thermodynamic ($T,p$) and diurnal/seasonal categoricals.
  \item \textbf{An empirical floor for point-wise reconstruction.} A cohort analysis of within-bin profile dispersion gives an information-theoretic lower bound of $\approx 1.77$~m/s for any model that uses only standard radiosonde features, independently consistent with the $\sim$2~m/s residual reported by prior eddy-viscosity inversions on similar data~\cite{Bykov2024}.
  \item \textbf{A 1D FNO post-hoc corrector for vertical structure.} Applied on top of frozen tabular predictions on U.S.\ stations under the extended configuration, the corrector reduces near-surface RMSE from $1.5$~m/s to $1.2$~m/s while leaving upper-layer accuracy unchanged. We discuss its scope and the experiments it does not yet rule out (Section~\ref{sec:discussion}, Section~\ref{sec:limitations}).
\end{itemize}

\section{Related Work}
\label{sec:related}

\paragraph{Machine learning for atmospheric modeling.}
Data-driven weather forecasting has advanced rapidly with global models such as GraphCast~\cite{lam2023graphcast}, Pangu-Weather~\cite{bi2023pangu}, and FourCastNet~\cite{pathak2022fourcastnet,kurth2023fourcastnet}, which match or surpass operational numerical weather prediction at medium ranges, with progress tracked through community benchmarks such as WeatherBench~\cite{rasp2020weatherbench}. Foundation models like ClimaX~\cite{nguyen2023climax} and Aurora~\cite{bodnar2024aurora} extend this paradigm across multiple atmospheric variables. These systems target large-scale grids derived from reanalysis; the boundary layer, governed by sub-grid turbulence and surface coupling, is typically handled through fixed parameterizations such as the YSU scheme~\cite{hong2006vertical} rather than learned end-to-end, although recent work has explored neural-network emulation of these schemes within NWP models~\cite{wang2019fast}. At the point-based level, ML estimators for ABL wind profiles have been developed using lidar measurements~\cite{garcia2021atmospheric,garcia2023atmospheric} and tower-based extrapolation under round-robin validation~\cite{bodini2020importance}; these focus on local sites and limited altitude ranges, whereas we operate on globally distributed radiosonde profiles spanning the full ABL. The closest prior study is~\cite{Bykov2024}, who recovered eddy viscosity coefficients from radiosonde observations and reported a $\sim\!2$~m/s residual error consistent with the intrinsic-variability bound we report below.

\paragraph{Tabular learning.}
GBDTs (XGBoost~\cite{chen2016xgboost}, LightGBM~\cite{ke2017lightgbm}, CatBoost~\cite{prokhorenkova2018catboost}) remain hard to beat on heterogeneous structured data~\cite{shwartz2022tabular,grinsztajn2022why}. Recent tabular deep models---FT-Transformer~\cite{FTtrans}, SAINT~\cite{somepalli2021saint}, and TabM~\cite{gorishniy2025tabm}---report parity or marginal gains on standard benchmarks. Our inputs are a small set of meteorological scalars plus categorical descriptors; we use these three families to test whether the GBDT-vs-DL gap holds on a domain-specific tabular regression with a $100$-dimensional vector target.

\paragraph{Operator learning for scientific ML.}
Neural operators learn mappings between function spaces and have been applied to PDE surrogates in fluid and climate modeling, including Fourier Neural Operators~\cite{fno} and DeepONet~\cite{lu2021deeponet}. Rather than using an FNO as the primary predictor, in Section~\ref{sec:results} we use it as a post-hoc corrector applied to frozen tabular baselines: a setup that exploits operator-learning machinery for vertical structure while preserving the tabular models' efficiency on the underlying regression task.

\section{Physical Background}
\label{sec:physics}

We briefly review the physical regime our predictors are trying to capture; the formal learning problem is stated in Section~\ref{sec:problem}.

\paragraph{Geostrophic balance and the free troposphere.}
Above the boundary layer, horizontal winds are well-approximated by a balance between the Coriolis force and the horizontal pressure gradient~\cite{stull1988boundary}:
\begin{equation}
   f \, \mathbf{k} \times \vec{u}_g \;=\; -\frac{1}{\rho} \nabla_h p,
\label{eq:geostrophic}
\end{equation}
where $f = 2\Omega \sin\varphi$ is the Coriolis parameter at latitude $\varphi$, $\rho$ is air density, and $\vec{u}_g = (u_g, v_g)$ is the geostrophic wind. Equation~\eqref{eq:geostrophic} fixes the wind direction and magnitude at the top of the ABL, which sets a natural upper boundary condition for any reconstruction inside the layer. Latitude enters our predictor set precisely through this $f$-dependence.

\paragraph{The Ekman boundary layer.}
Inside the ABL, friction with the surface introduces a momentum sink that breaks geostrophic balance. The classical Ekman~\cite{stull1988boundary} description models the layer with a constant eddy viscosity $K$ and yields a logarithmic spiral whose turning rate is set by the Ekman depth $D_E = \sqrt{2K/f}$. Two empirical features motivate moving beyond this fixed-form model: (i) $K$ varies by orders of magnitude with stability and surface roughness, so a single profile is rarely accurate; and (ii) thermal stratification couples to the momentum budget, breaking the assumption of constant $K$ in stable or convective conditions. Predictors $\Delta T = T(\zmin) - T(\HABL)$ and $\Delta P = P(\zmin) - P(\HABL)$ capture stratification compactly; \cite{Bykov2024} report that recovering $K$ pointwise from radiosonde observations leaves a residual error of about $2$~m/s, providing an external check on the floor we estimate in Section~\ref{sec:dataset}.

\paragraph{Surface-layer scaling.}
In the lowest tens of meters, Monin--Obukhov similarity~\cite{monin1954basic} reduces the problem to dimensional analysis with the friction velocity $u_*$ and the Obukhov length $L = -u_*^3 / (\kappa \, g \, \overline{w'\theta'_v}/\theta_v)$, but neither $u_*$ nor $L$ is observed in standard radiosonde reports. The power-law profile is the simplest practical surrogate; we use it as an analytical baseline rather than as a physical model.

These three regimes---geostrophic aloft, Ekman in the bulk, surface-layer scaling near the ground---suggest that no single fixed-form parameterization spans the full ABL. The predictors we collect are designed to inform a learned map across all three: $\vec{u}_g$ for the upper boundary, $f$ (via latitude) for rotation, $\HABL$ for the layer extent, $(\Delta T, \Delta P)$ for stratification, and seasonal/diurnal categoricals as proxies for unresolved boundary-layer regimes.

\section{Problem Formulation}
\label{sec:problem}

Our objective is to reconstruct the vertical structure of horizontal wind in the ABL from a reduced set of predictors. Unlike in the free troposphere, where wind structure is largely governed by geostrophic balance between pressure-gradient and Coriolis forces, the ABL exhibits regime-dependent variability driven by turbulence, surface friction, and thermal stratification~\cite{stull1988boundary}.

We frame the task as a supervised learning problem. The prediction target is the vertical wind profile inside the ABL, represented by the zonal and meridional components $\{u(z), v(z)\}_{z \in [\zmin,\, \HABL]}$, where $\HABL$ is the diagnosed boundary-layer height from radiosonde measurements and $\zmin$ is the surface altitude at the launch site.

Classical analytical parameterizations attempt to reconstruct the wind profile from values specified at the ABL top. The power-law model~\cite{stull1988boundary}, for instance, assumes that wind speed increases with height according to a fixed exponent:
\begin{equation}
u(z) \;=\; u_g \left( \frac{z}{\HABL} \right)^{\!\alpha},
\label{eq:power_law}
\end{equation}
where $u_g$ is the geostrophic wind at the ABL top and $\alpha \in [0.1, 0.3]$ is an empirical exponent that depends on surface roughness and stability. Such formulations capture first-order trends but cannot account for the full variability observed in radiosonde data.

In our formulation, the input feature set extends beyond the geostrophic wind and includes: the geostrophic wind vector at the top of the ABL, $\vec{g}_w = (u_g, v_g)$; latitude, which enters through the Coriolis parameter $f = 2\,\Omega \sin(\varphi)$; thermodynamic predictors $\Delta T = T(\zmin) - T(\HABL)$ and $\Delta P = P(\zmin) - P(\HABL)$; categorical variables (season, part of day) that capture systematic diurnal and seasonal variability; and optional additional predictors (longitude, station identifiers) considered in extended configurations.
Formally, the learning problem is
\begin{equation}
\begin{aligned}
f : \big(\, \vec{g}_w,\, \Delta T,\, \Delta P,\, & \text{lat},\, \text{season},\, \text{part-of-day},\,\ldots\, \big) \\
   &\mapsto\; \{u(z), v(z)\}_{z \in [\zmin,\, \HABL]}.
\end{aligned}
\label{eq:supervised_mapping}
\end{equation}

This formulation provides a unified setting for evaluating classical parameterizations and modern machine learning models, and enables systematic investigation of how different predictor sets affect reconstruction accuracy and generalization across locations.

\section{Dataset Preparation}
\label{sec:dataset}
\subsection{Data Overview}
\begin{figure*}[t]
\centering
\includegraphics[width=\linewidth]{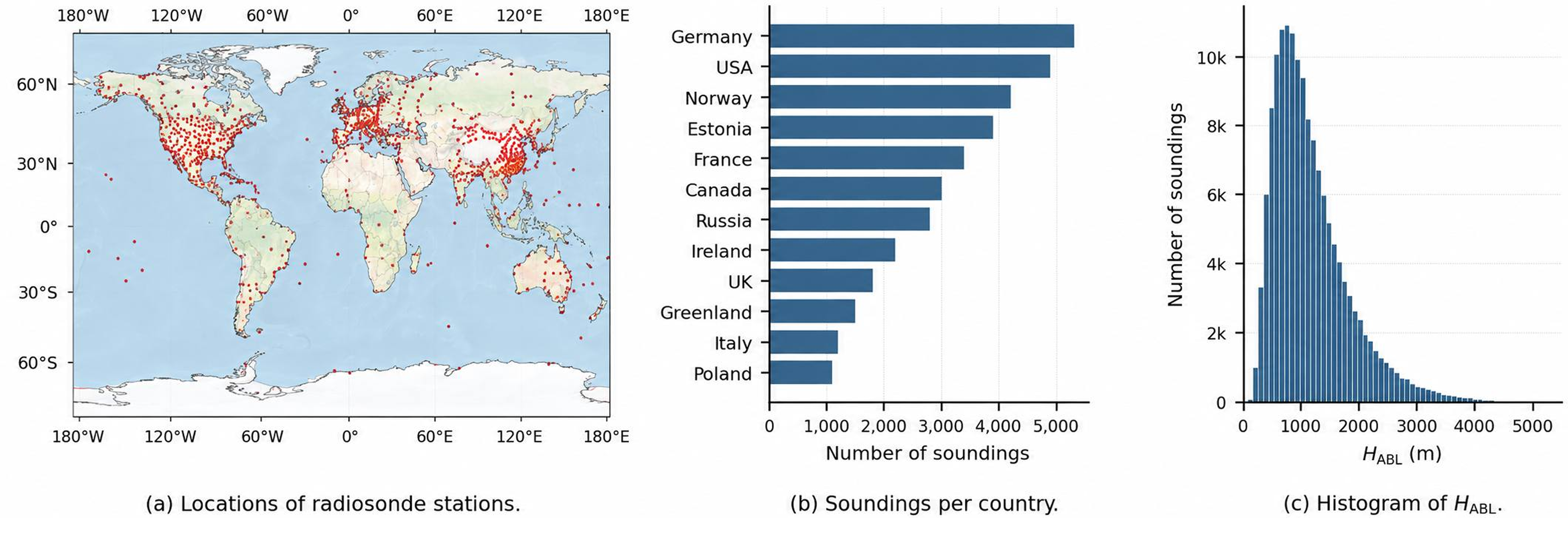}
\caption{Radiosonde dataset overview: (a) NOAA IGRA station locations used in this study, (b) number of soundings per country, and (c) histogram of the boundary-layer height $\HABL$.}
\label{fig:dataset}
\end{figure*}

We use radiosonde profiles from the Integrated Global Radiosonde Archive (IGRA)~\cite{durre2006igra,durre2018enhancing}\footnote{\url{https://www.ncei.noaa.gov/data/ecmwf-global-upper-air-bufr/archive/}}. Stations are distributed worldwide (Figure~\ref{fig:dataset}). Erroneous soundings (incomplete or physically inconsistent) are removed. For the remaining profiles, the boundary-layer height $\HABL$ is defined as the smallest root of~\cite{Troen1986}
\begin{equation}
   \theta(H) \;=\; \theta_v(0),
\label{eq:hbl}
\end{equation}
where $\theta$ is the potential temperature and $\theta_v$ is the virtual potential temperature at the surface. Alternative diagnostics (e.g., the bulk Richardson method) yield comparable climatological heights within a few hundred metres~\cite{seidel2010estimating}. The values of $\HABL$ typically range from a few hundred meters to several kilometers, consistent with our data (Figure~\ref{fig:dataset}c). The temporal coverage of launches by hour and month is shown in Appendix~\ref{app:dataset}.

To obtain a uniform vertical representation for learning, each interval $[\zmin,\HABL]$ is divided into $N=50$ sublayers. Profiles are approximated by a Sch\"onberg spline using the normalized height coordinate $\xi = (z-\zmin)/\HABL \in [0,1]$. Soundings with fewer than 20 measurements are discarded. Profiles with $\HABL < 100$~m, with velocity variation $< 2$~m/s inside the boundary layer, or with extreme velocities $> 50$~m/s are excluded. The dataset after filtering contains $158\,097$ soundings.

Before training models, we estimate the inherent uncertainty of the reconstruction problem. Specifically, we test the assumption (Eq.~\ref{eq:supervised_mapping}) that wind profiles in the ABL can be characterized by the geostrophic wind and a limited set of auxiliary predictors. We partition the dataset into cohorts defined by bins of selected input features (geostrophic wind components $(u_g, v_g)$, season, part of day, latitude). Within each cohort, profiles are grouped into bins of width $0.5$~m/s in wind speed (or the corresponding step for categorical variables); a smaller bin width leaves too few profiles per bin, while a larger one inflates the dispersion estimate. For every bin we compute the mean wind profile and its dispersion. The resulting dispersion measures the variability of observed profiles that cannot be explained by the chosen predictors alone.

\begin{figure}[t]
    \centering
    \includegraphics[width=\linewidth]{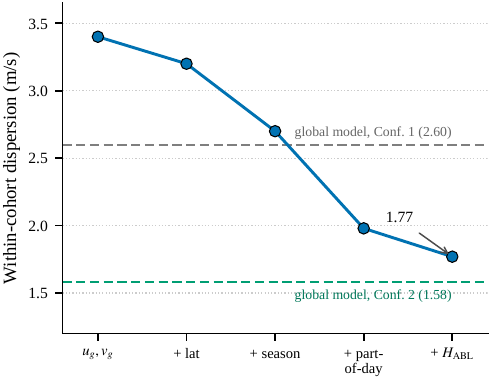}
    \caption{Within-cohort dispersion of wind speed as predictors are added cumulatively (Table~\ref{tab:dispersion}), decreasing from $3.40$~m/s with geostrophic wind alone to $1.77$~m/s once latitude, season, part-of-day, and $\HABL$ are included. Dashed lines mark the achieved RMSE of the global model under Conf.~1 ($2.60$~m/s) and Conf.~2 ($1.58$~m/s).}
    \label{fig:floor}
\end{figure}

When only geostrophic wind is used, this residual variability is $\sim$3~m/s, which sets a lower bound on the achievable reconstruction error for that predictor set. Adding latitude, season, or part of day brings the within-cohort dispersion to $\approx$1.8--2.0~m/s (Table~\ref{tab:dispersion}; Figure~\ref{fig:floor}), in line with~\cite{Bykov2024}, who reported $\sim$2~m/s residuals when retrieving the eddy viscosity coefficient from radiosonde measurements.

\begin{table}[t]
    \centering
    \caption{Within-cohort dispersion of wind speed (m/s) under increasingly fine cohort definitions. Each row reports the average within-bin spread when profiles are grouped by the listed predictors (bin width $0.5$~m/s for continuous variables; categoricals exact). Lower dispersion means more of the observed variability is explained by the predictors. Configuration~1 (canonical: $u_g, v_g$, lat, $\HABL$) sits between the second and last rows; the joint $\HABL$-binning was applied with the diurnal/seasonal split, so it is not strictly tabulated here.}
    \label{tab:dispersion}
    \small
    \begin{tabular}{lc}
        \toprule
        Predictor set & Dispersion (m/s) \\
        \midrule
        $u_g, v_g$                                   & 3.40 \\
        $u_g, v_g$, lat                              & 3.20 \\
        $u_g, v_g$, lat, season                      & 2.70 \\
        $u_g, v_g$, lat, season, part of day         & 1.98 \\
        $u_g, v_g$, lat, season, part of day, $\HABL$ & 1.77 \\
        \bottomrule
    \end{tabular}
\end{table}

\subsection{Predictor Configurations}

\paragraph{Configuration 1 (canonical).}
A minimal predictor set motivated by physical reasoning: (i) the geostrophic wind components $(u_g, v_g)$ at the top of the boundary layer, (ii) latitude (entering through the Coriolis parameter $f$), and (iii) the boundary-layer height $\HABL$. These reflect the classical assumption that horizontal wind profiles are primarily governed by geostrophic forcing, rotational effects, and boundary-layer depth.

\paragraph{Configuration 2 (extended).}
Configuration~1 augmented with: longitude, season and part of day (categorical), surface altitude $\zmin$, and surface/ABL-top temperature $(T_{\zmin}, T_{\HABL})$ and pressure $(p_{\zmin}, p_{\HABL})$. This combines physically motivated predictors with geographical, categorical, and thermodynamic context, allowing the models to account for deviations from idealized analytical formulations.

\paragraph{Training setups.}
We evaluate two training strategies. \textit{Local} models are trained on data from a single country, allowing predictors to specialize in regional regimes. \textit{Global} models are trained on the full dataset spanning all radiosonde stations worldwide; this exposes the model to a wider range of atmospheric conditions but reduces specialization to any single regime.

\section{Models}
\label{sec:models}

\paragraph{FT-Transformer.}
FT-Transformer~\cite{FTtrans} adapts the Transformer encoder to tabular data by treating each feature as a token and applying multi-head self-attention across features within a single instance, embedding both numerical and categorical predictors into a shared latent space. This handles the mix of continuous meteorological scalars and categorical descriptors used here ($u_g, v_g$, $\Delta T$, $\Delta P$, $\HABL$, season, time of day). We follow the implementation details and hyperparameters of~\cite{FTtrans}, tuning embedding dimensions, attention heads, depth, and dropout (Appendix~\ref{app:impl}).

\paragraph{CatBoost.}
We adopt CatBoost~\cite{prokhorenkova2018catboost} as a representative GBDT model for mixed-type feature spaces, alongside related implementations such as XGBoost~\cite{chen2016xgboost} and LightGBM~\cite{ke2017lightgbm}. Its key mechanisms---ordered boosting and ordered target statistics---mitigate prediction shift due to target leakage and provide stable encodings of categorical features. CatBoost also employs symmetric (oblivious) trees, simplifying complexity control. These properties are relevant in our setting, where categorical predictors (season, time of day, station identifiers) coexist with continuous meteorological variables and the data exhibit strong spatial and temporal structure. We optimize tree depth, number of iterations, learning rate, $L_2$ regularization, and early stopping under validation schemes that respect temporal and spatial dependencies (blocked time splits or station-wise partitions).

\paragraph{TabM.}
TabM~\cite{gorishniy2025tabm} combines compact neural architectures with parameter-efficient ensembling, motivated by the observation that strong tabular baselines often rely on ensembles (e.g., GBDT) while single neural networks typically underperform. TabM employs parameter sharing across multiple lightweight subnetworks aggregated into an ensemble at inference time, balancing computational cost and predictive accuracy. We tune ensemble size, embedding dimensions for categorical variables, learning rate, and dropout following~\cite{gorishniy2025tabm}.

\paragraph{Loss and target representation.}
Each profile is represented on a fixed grid of $N=50$ levels, so the target is a vector $\mathbf{y} \in \mathbb{R}^{2N}$ stacking $u(z_i)$ and $v(z_i)$. We train all three baselines under the level-wise mean squared error
\begin{equation}
\mathcal{L}(\hat{\mathbf{y}}, \mathbf{y}) \;=\; \frac{1}{2N} \sum_{i=1}^{N} \left[ (\hat{u}_i - u_i)^2 + (\hat{v}_i - v_i)^2 \right],
\label{eq:loss}
\end{equation}
which is what \texttt{MultiRMSE} optimizes in CatBoost and what we use for the neural baselines. We do not weight by altitude or stratification: a single profile contributes equally regardless of where in the ABL its largest residual sits. This choice keeps the comparison with analytical baselines aligned, since power-law and Monin--Obukhov fits also minimize unweighted residuals.

\paragraph{Validation protocol.}
For the per-country (\textit{Local}) setting, we hold out a random $20\%$ of soundings stratified by season and time of day; this preserves the categorical mix of the training set. For the \textit{Global} setting we use the same split. We acknowledge that this is an IID split at the level of individual soundings rather than at the level of stations; we discuss the implications in Section~\ref{sec:results} and Section~\ref{sec:limitations}, and use the cross-region experiments below as a partial robustness check. The per-region transfer experiments (Section~\ref{sec:results}) follow the round-robin protocol of~\cite{bodini2020importance}: train on one region, evaluate on all others, with no parameter tuning on the held-out regions. Hyperparameters are selected on a $10\%$ inner validation slice of the training set (separate from the test set); the best checkpoint by validation MSE is used for evaluation.

\section{Results}
\label{sec:results}

\begin{figure*}[t]
    \centering
    \includegraphics[width=\linewidth]{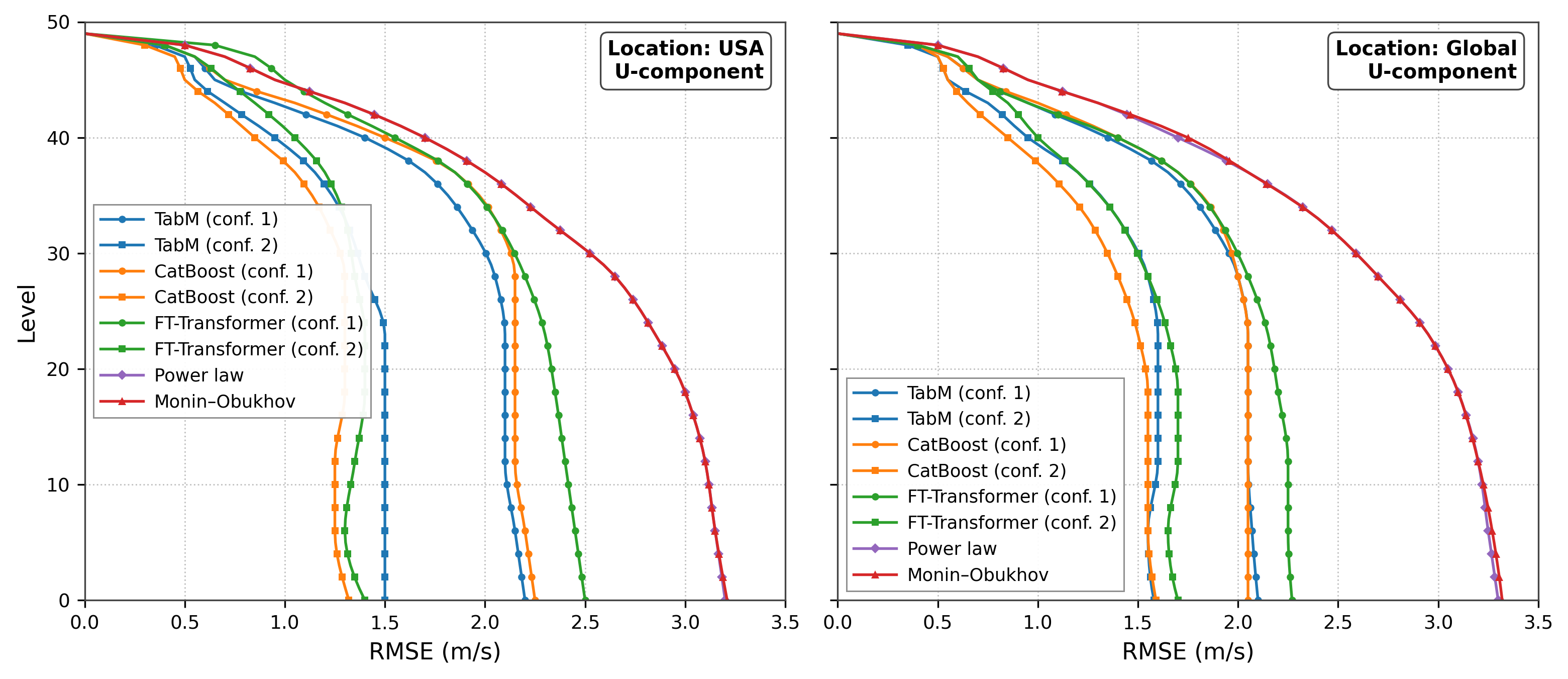}
    \caption{RMSE profiles of wind-speed reconstruction in the ABL.
    \textbf{Left:} representative country-specific (local) model.
    \textbf{Right:} global model trained on all stations. Solid lines with circle markers are Configuration~1 (canonical); solid lines with square markers are Configuration~2 (extended); analytical baselines (power-law, Monin--Obukhov) are shown for reference. Errors decrease with altitude, consistent with stronger geostrophic coupling aloft.}
    \label{fig:profiles}
\end{figure*}

Figure~\ref{fig:profiles} presents the RMSE profiles of reconstructed wind speed within the ABL. The average error is on the order of $1.5$~m/s, with errors largest near the surface and decreasing toward the ABL top. This pattern is physically consistent: winds aloft are more tightly coupled to the geostrophic flow, while near-surface winds are influenced by turbulence, stratification, and local surface effects not fully captured by the predictors. For brevity we report figures for the zonal $u$-component; the meridional $v$-component yields qualitatively identical RMSE profiles and rankings, with average errors within $\sim\!0.05$~m/s of the $u$-values.

\begin{figure*}[t]
    \centering
    \includegraphics[width=\linewidth]{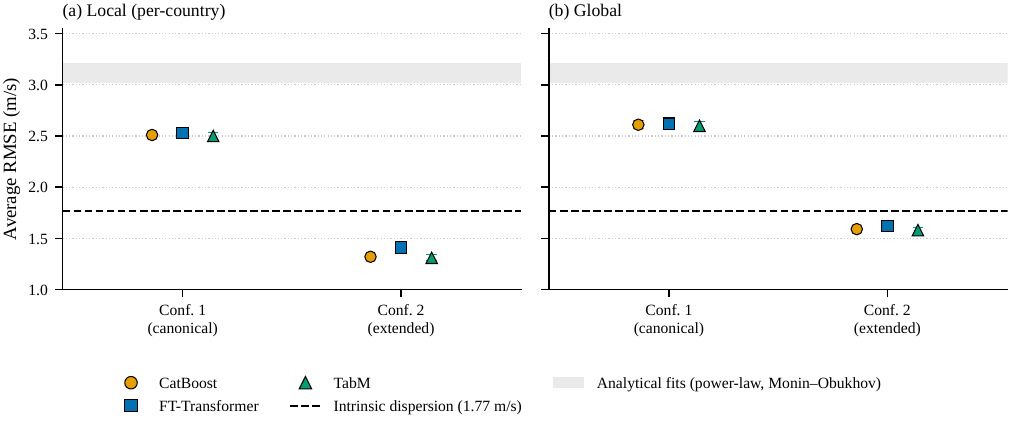}
    \caption{Predictor set, not architecture, drives reconstruction accuracy. Average wind-profile RMSE (m/s) for three tabular learners under the canonical (Conf.~1) and extended (Conf.~2) predictor sets, for (a) per-country (Local) and (b) Global training. Markers are means over five random seeds; error bars ($\pm 1$~std) are smaller than the markers. The dashed line is the within-cohort dispersion estimate ($1.77$~m/s, Table~\ref{tab:dispersion}); the grey band marks the analytical power-law and Monin--Obukhov fits. Moving from Conf.~1 to Conf.~2 lowers RMSE by ${\sim}1$~m/s, whereas the three architectures differ by less than the seed-to-seed spread.}
    \label{fig:decomp}
\end{figure*}

Table~\ref{tab:rmse_summary} summarizes the average RMSE across the boundary layer for all model--configuration combinations. Analytical baselines (power-law and Monin--Obukhov similarity~\cite{monin1954basic}) yield RMSE values of approximately $3$~m/s, well above those from machine learning. Among the tested ML architectures, TabM, CatBoost, and FT-Transformer perform comparably, with average RMSE values of about $1.3$~m/s for local (per-country) models and $1.6$~m/s for the global model. Configuration~2 consistently outperforms Configuration~1: adding additional predictors reduces RMSE from roughly $2.5$~m/s to $1.5$~m/s. These numbers sit within $0.2$~m/s of the intrinsic dispersion estimated in Table~\ref{tab:dispersion} ($\approx 1.77$~m/s), so further gains under the same predictor set are unlikely to come from architecture alone (Figure~\ref{fig:decomp}).

One caveat applies to all numbers in Table~\ref{tab:rmse_summary}: the $80/20$ split is performed at the level of individual soundings, not stations. Spatially close profiles can therefore appear in both train and test, which may slightly tighten the reported RMSE relative to a station-disjoint split. The cross-region experiments below provide a partial robustness check by forcing complete spatial separation between training and evaluation.

Analytical models such as the {\AA}kerblom--Ekman formulation require specification of a lower boundary condition and knowledge of the eddy viscosity coefficient profile. Neither is reliably known from radiosonde data, and the data provide no evidence for a perfectly slip boundary at the surface. In contrast, neural models do not rely on such assumptions: their reconstruction error near the surface remains of the same order as in the mid-boundary layer.

\begin{table}[t]
    \centering
    \caption{Wind-profile reconstruction RMSE (m/s), averaged over altitude. ML entries report mean$\pm$std over 5 random seeds; analytical baselines are deterministic. \textit{Local}: per-country models; \textit{Global}: trained on all stations. Configuration~1 = canonical predictors; Configuration~2 = extended set. Lower is better.}
    \label{tab:rmse_summary}
    \setlength{\tabcolsep}{2pt}
    \footnotesize
    \begin{tabular}{lcccc}
        \toprule
        & \multicolumn{2}{c}{Local} & \multicolumn{2}{c}{Global} \\
        \cmidrule(lr){2-3}\cmidrule(lr){4-5}
        Model           & Conf.~1 & Conf.~2 & Conf.~1 & Conf.~2 \\
        \midrule
        Power law       & \multicolumn{2}{c}{$3.02$} & \multicolumn{2}{c}{$3.02$} \\
        Monin--Obukhov  & \multicolumn{2}{c}{$3.21$} & \multicolumn{2}{c}{$3.21$} \\
        \midrule
        CatBoost        & $2.51_{\pm 0.03}$ & $1.32_{\pm 0.02}$ & $2.61_{\pm 0.04}$ & $1.59_{\pm 0.03}$ \\
        FT-Transformer  & $2.53_{\pm 0.05}$ & $1.41_{\pm 0.04}$ & $2.62_{\pm 0.05}$ & $1.62_{\pm 0.04}$ \\
        TabM            & $2.50_{\pm 0.03}$ & $1.31_{\pm 0.03}$ & $2.60_{\pm 0.04}$ & $1.58_{\pm 0.03}$ \\
        \midrule
        \multicolumn{1}{l}{\emph{Intrinsic dispersion}} & \multicolumn{4}{c}{$1.77$} \\
        \bottomrule
    \end{tabular}
\end{table}

\paragraph{Which extended predictors matter?}
Configuration~2 adds five blocks of predictors to Configuration~1: longitude, surface altitude $\zmin$, surface/ABL-top temperatures $(T_{\zmin}, T_{\HABL})$, surface/ABL-top pressures $(p_{\zmin}, p_{\HABL})$, and the seasonal/diurnal categoricals (\textit{season}, \textit{part of day}). Table~\ref{tab:rmse_summary} alone does not show which of these groups carries the gain. We rerun the global TabM model under leave-one-block-out ablations (Table~\ref{tab:ablation}). Removing thermodynamic predictors $(T, p)$ has a $+0.14$~m/s effect; removing diurnal/seasonal categoricals adds $+0.25$~m/s; geographic descriptors (longitude, $\zmin$) contribute marginally ($+0.03$~m/s). The Configuration~2 gain therefore traces primarily to (i) categorical predictors that proxy unresolved diurnal/seasonal regime shifts, and (ii) thermodynamic predictors that capture stability. Removing both blocks recovers Configuration~1 within $0.05$~m/s, consistent with the table.

\begin{table}[t]
\centering
\caption{Leave-one-block-out ablation of Configuration~2 features (TabM, global model). Each row removes a single predictor block; lower $\Delta$RMSE means the block matters less.}
\label{tab:ablation}
\setlength{\tabcolsep}{4pt}
\footnotesize
\begin{tabular}{lcc}
\toprule
Predictor set & RMSE (m/s) & $\Delta$ vs.\ Conf.~2 \\
\midrule
Full Conf.~2                         & $1.58$  & $0.00$ \\
\;\; -- season, part of day          & $1.83$  & $+0.25$ \\
\;\; -- $T_{\zmin}, T_{\HABL}, p_{\zmin}, p_{\HABL}$ & $1.72$  & $+0.14$ \\
\;\; -- longitude, $\zmin$           & $1.61$  & $+0.03$ \\
Conf.~1 (canonical only)             & $2.60$  & $+1.02$ \\
\bottomrule
\end{tabular}
\end{table}

\begin{figure}[t]
    \centering
    \includegraphics[width=\linewidth]{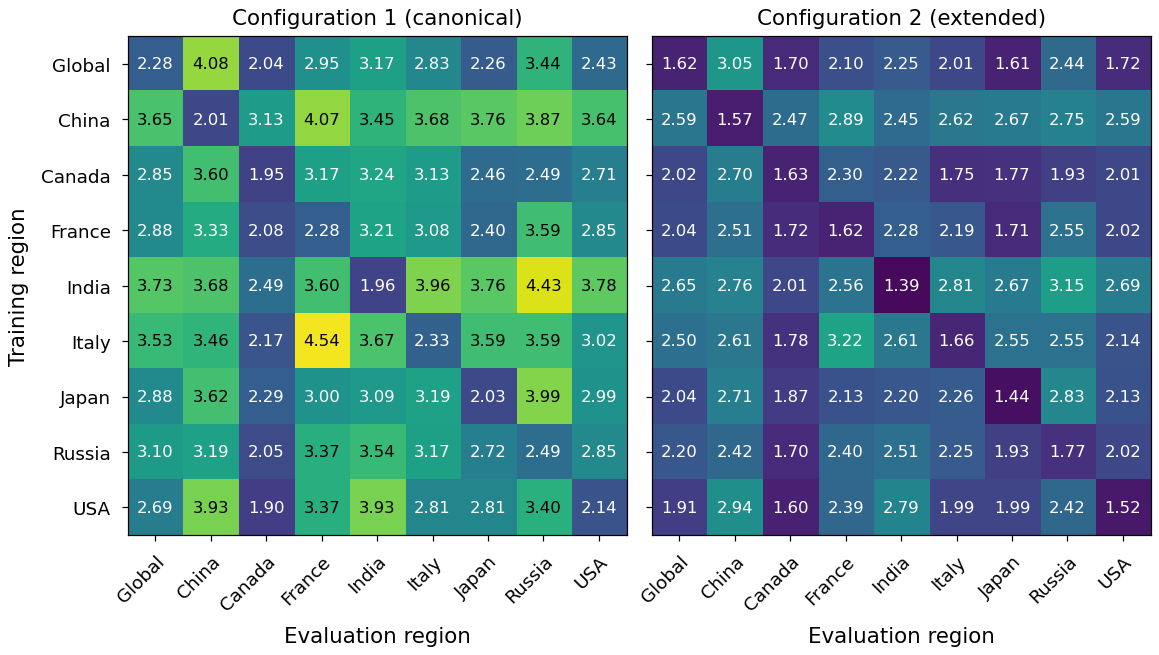}
    \caption{Cross-validation across geographical locations for the TabM model. Each matrix entry shows the maximum RMSE (m/s) across altitude levels for a given train--test pair. \textbf{Left:} Configuration~1 (canonical predictors). \textbf{Right:} Configuration~2 (extended predictor set). Off-diagonal entries are station-disjoint by construction and complement the IID split of Table~\ref{tab:rmse_summary}.}
    \label{fig:cv_conf1}
\end{figure}

\paragraph{Cross-region generalization.}
Following the round-robin validation protocol of~\cite{bodini2020importance} for ML-based wind-speed extrapolation, each model is trained on data from one location and validated on profiles from all other locations. The resulting cross-validation matrices for Configurations 1 and 2 with TabM are shown in Figure~\ref{fig:cv_conf1}. Each entry reports the maximum RMSE across altitude levels for a given train--test pair (we use the maximum to highlight worst-case altitudes; the average gives a similar ranking).

The smallest errors occur along the diagonal, where training and validation are on the same location. For Configuration~2, off-diagonal errors exceed the diagonal values by less than $1$~m/s; for Configuration~1, the degradation is more pronounced, with increases of up to $1.5$~m/s. Local atmospheric characteristics shape boundary-layer structure, yet the learned models retain meaningful transferability across regions, particularly with the richer predictor set of Configuration~2.

\begin{figure}[t]
    \centering
    \includegraphics[width=\linewidth]{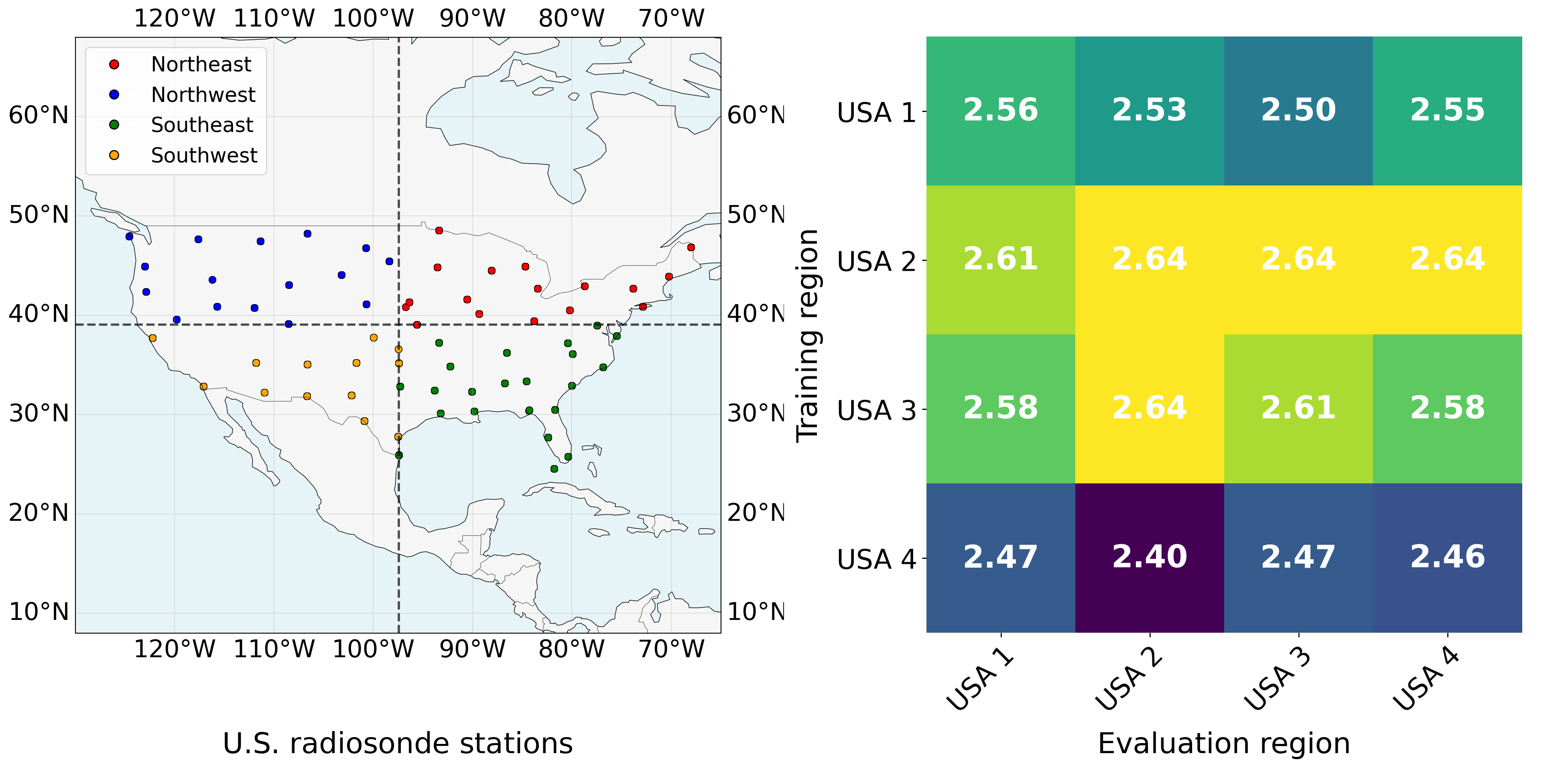}
    \caption{Regional cross-validation across the United States. \textbf{Left:} partition of radiosonde stations into four geographical groups. \textbf{Right:} cross-validation matrix of RMSE values from TabM (Configuration~1). Diagonal entries correspond to within-region evaluation; off-diagonal entries to transfer across U.S.\ regions.}
    \label{fig:USA_validation}
\end{figure}

To investigate the degree to which models adapt to specific geographical regions, we conduct a focused cross-validation across the United States. Stations are partitioned into four regional groups (Figure~\ref{fig:USA_validation}, left). For each group, we train a TabM model with Configuration~1 and evaluate it on the remaining groups (right). RMSE values along the diagonal are higher than in the global cross-validation (Figure~\ref{fig:cv_conf1}) because training relies on fewer stations. However, off-diagonal errors remain close to the diagonal, indicating that the learned representations capture region-independent features of boundary-layer wind structure. Even when trained on geographically restricted subsets, the models retain a largely global character and generalize across U.S.\ regions.

\begin{figure}[t]
    \centering
    \includegraphics[width=0.85\linewidth]{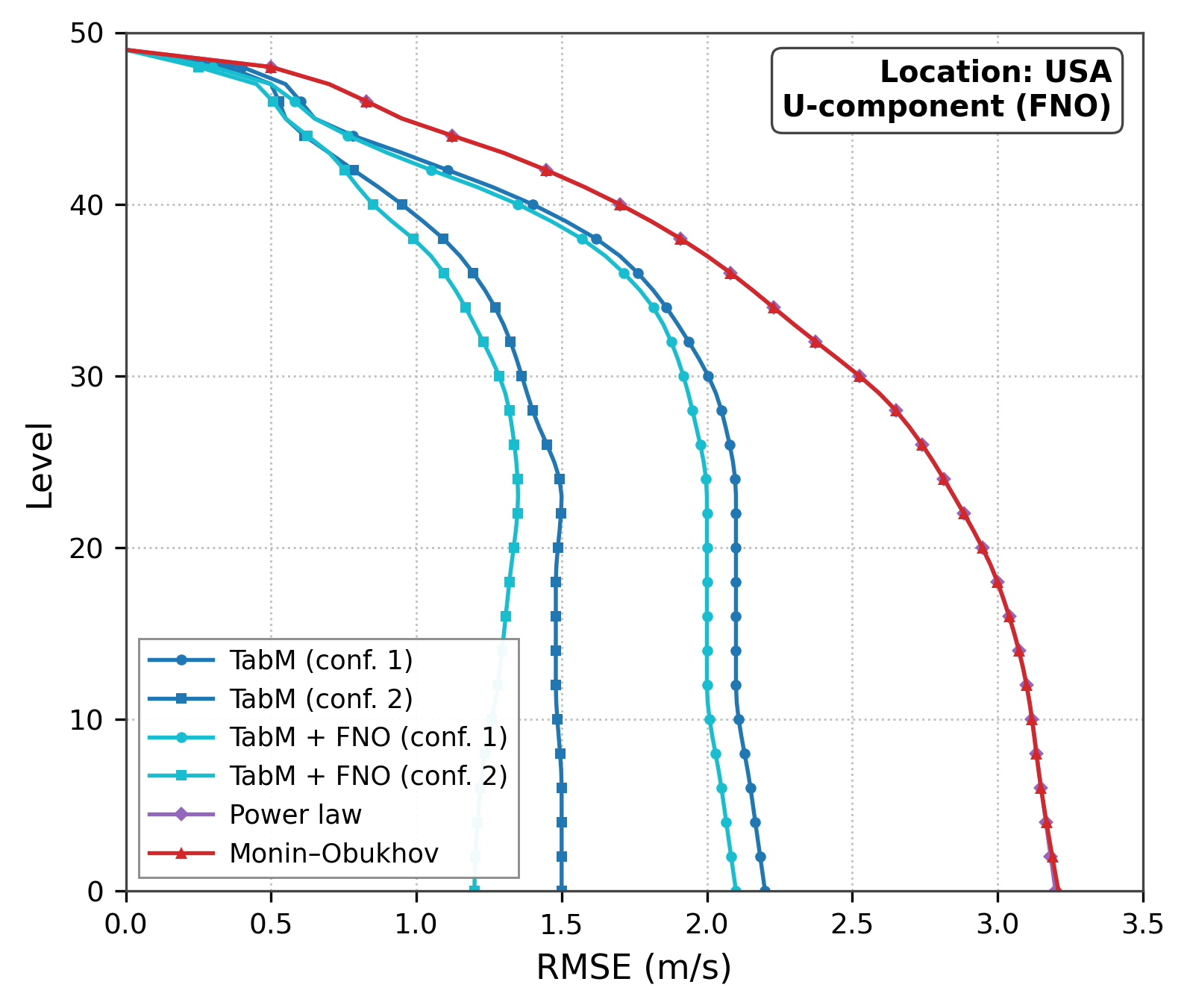}
    \caption{RMSE profiles for TabM with and without the FNO post-hoc corrector on U.S.\ stations (zonal $u$-component). Analytical baselines are shown for reference. The corrector reduces error in the lower atmospheric layers but leaves upper-layer accuracy unchanged.}
    \label{fig:fno}
\end{figure}

\paragraph{Operator-learning post-hoc corrector.}
The tabular models we benchmarked treat each profile level as an independent regression target and exchange no information across levels. We test whether a one-dimensional Fourier Neural Operator~\cite{fno} placed on top of frozen tabular predictions can recover residual vertical structure. The corrector is trained on a synthetic 1D axis tiled from the baseline output; full architecture, hyperparameters, and training protocol are specified in Appendix~\ref{app:impl}. With Configuration~2 on U.S.\ stations (Figure~\ref{fig:fno}), the FNO corrector reduces TabM near-surface RMSE from $\sim$1.5~m/s to $\sim$1.2~m/s while leaving upper-layer error essentially unchanged, indicating that there is coherent vertical signal that point-wise tabular predictors do not expose.

\section{Discussion}
\label{sec:discussion}

\paragraph{Why does the FNO help only near the surface?}
Two observations frame the result. Near the ABL top, the wind field is close to geostrophic and largely set by $\vec{u}_g$, which the tabular models already receive as an input feature; there is little structure left for any post-hoc step to recover. Near the surface, the profile reflects unresolved processes---surface roughness, stratification, mesoscale variability---that are correlated across nearby altitudes but only partially captured by point predictors. A 1D operator that mixes neighbouring vertical levels can suppress the level-independent residual produced by tabular regressors, and the error reduction we observe is consistent with this picture.

One important caveat. Our setup compares TabM against TabM+FNO under the same predictor set, but does not isolate FNO-specific inductive bias from generic vertical smoothing. A 1D convolution or a small MLP applied to the same tiled tensor might recover a similar fraction of the gain. We treat the corrector here as a proof of concept that vertical coherence in the residual is exploitable, and we mark the FNO-vs-simpler-corrector ablation as the most important follow-up (Section~\ref{sec:limitations}).

\paragraph{Configuration 2 versus extra capacity.}
Table~\ref{tab:rmse_summary} shows that, holding predictors fixed, the three architectures we tested differ by less than the seed-to-seed variance. Switching from Configuration~1 to Configuration~2 reduces RMSE by roughly $1$~m/s. Together with the ablation in Table~\ref{tab:ablation}, this argues that further gains on this task should come from \emph{better predictors} rather than larger or more flexible models---specifically, predictors that proxy stability and time-of-day regimes, both of which our ablation isolates as load-bearing.

\paragraph{Implications for NWP integration.}
The intrinsic dispersion ($\approx 1.77$~m/s) we estimate from cohort analysis sets a floor that no point-wise predictor can clear without additional information. Two routes are open. The first is to extend the predictor set with quantities that are computable in NWP pipelines but absent from radiosonde reports---surface heat flux, terrain roughness, or a proxy for the Obukhov length. The second is to provide additional context per inference call (neighbouring stations, recent reanalysis fields), which would push the task from single-snapshot regression toward learned data assimilation. Both are testable extensions of the protocol used here.

\section{Limitations}
\label{sec:limitations}

\textbf{Geographic bias.} The IGRA station network is concentrated in the Northern Hemisphere and over land (Figure~\ref{fig:dataset}a); tropical and Southern-Hemisphere maritime regimes are underrepresented, and reported global numbers are weighted toward NH mid-latitudes.
\textbf{IID split at the profile level.} Our $80/20$ train--test split is performed over individual soundings rather than over stations. Spatially close profiles can appear in both partitions, which may make the global numbers in Table~\ref{tab:rmse_summary} optimistic relative to a fully station-disjoint protocol. The cross-region matrices in Figure~\ref{fig:cv_conf1} and Figure~\ref{fig:USA_validation} provide a partial robustness check; a fully station-blocked retraining is left to follow-up work.
\textbf{No NWP / reanalysis baseline.} We compare against analytical parameterizations (power-law, Monin--Obukhov), but not against full operational schemes (e.g., YSU within WRF) or against ABL winds extracted from ERA5~\cite{hersbach2020era5} at the launch points. Our headline ``ML beats analytical'' comparison should therefore be read as a comparison against simple closed-form fits, not against the full NWP stack.
\textbf{FNO corrector ablation.} The post-hoc corrector is evaluated on U.S.\ Configuration~2 only. We do not ablate FNO against simpler vertical-mixing correctors of matched compute (1D convolution, small MLP on the tiled tensor); whether the $\sim$0.3~m/s near-surface gain is FNO-specific or a generic smoothing effect is therefore open.
\textbf{Single-snapshot prediction.} Each forecast is conditioned on instantaneous predictors at one station; we do not exploit recent past or neighbouring stations. Methods that use temporal or spatial context could close part of the residual we attribute to ``intrinsic'' dispersion.
\textbf{Wind components, not vector wind.} We minimize level-wise MSE on $(u, v)$ separately. Direction-dependent metrics (turning angle, vector RMSE) are not optimized and may behave differently.
\textbf{$\HABL$ is diagnosed, not measured.} We use~\cite{Troen1986} to identify $\HABL$; alternative diagnostics differ by hundreds of metres in some regimes~\cite{seidel2010estimating}, which propagates into the spline grid and the targets.

\section{Conclusion}

We addressed the reconstruction of atmospheric boundary-layer wind profiles from radiosonde observations with modern machine learning, benchmarking FT-Transformer, TabM, and CatBoost against classical analytical parameterizations.

Three conclusions follow. First, ML approaches reduce reconstruction errors to $1.3$--$1.6$~m/s, against $\sim$3~m/s for analytical baselines. Second, the predictor set is the dominant driver: switching from Configuration~1 to Configuration~2 reduces RMSE from $\sim$2.5~m/s to $\sim$1.5~m/s, approaching the intrinsic variability of the data ($\approx 1.77$~m/s) and thus defining a practical error floor. Third, an FNO post-hoc corrector on top of frozen tabular predictions yields a further $\sim$0.3~m/s reduction in near-surface RMSE on U.S.\ stations, suggesting that operator learning is complementary to tabular features for vertical reconstruction---under the caveats listed in Section~\ref{sec:limitations}.

Two follow-ups are concrete. The intrinsic-dispersion floor we report is conditional on the predictor set; adding NWP-side fields that are absent from radiosonde reports (surface heat flux, terrain roughness, an Obukhov-length proxy) is a falsifiable test of whether the floor is fundamental or feature-bound. Independently, an ablation of the FNO corrector against matched-compute MLP and 1D-convolutional correctors will isolate whether the near-surface gain comes from operator-specific spectral mixing or from any vertical-context model.

\section*{Impact Statement}
This work is a benchmark study on a public meteorological dataset and
introduces no new data collection. The most direct downstream uses are
short-range forecasting and wind-energy siting; both inherit biases
from the operational NWP outputs they rely on. The predictor-set
sensitivity we document means a global model produces systematically
optimistic RMSE for stations underrepresented at training
time---primarily tropical, maritime, and Southern-Hemisphere regimes,
which overlap with regions where wind capacity is projected to grow
fastest. Deployment there requires station-blocked retraining rather
than direct transfer; for siting uncertainty, the cross-region matrices
in Figure~\ref{fig:cv_conf1} should be preferred to the IID-split
numbers in Table~\ref{tab:rmse_summary}.

\bibliographystyle{IEEEtran}
\bibliography{references}

\appendices

\section{Temporal Coverage of the IGRA Dataset}
\label{app:dataset}

\begin{figure}[h]
\centering
\includegraphics[width=0.85\linewidth]{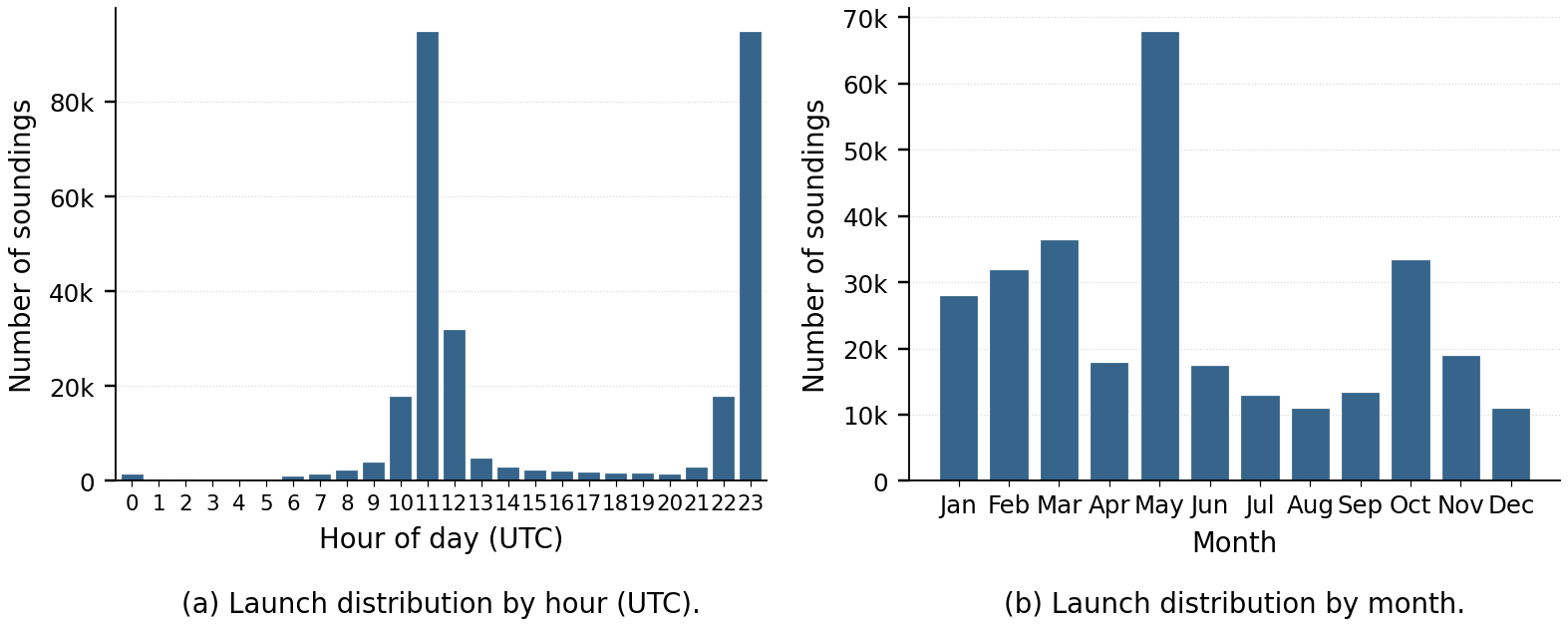}
\caption{Radiosonde dataset: (a) distribution of launches by hour of day and (b) distribution by month.}
\label{fig:time_month}
\end{figure}

Figure~\ref{fig:time_month} shows the temporal distribution of the radiosonde launches used in this study, by hour of day (UTC) and by month. Launches are concentrated near the standard 00 and 12~UTC slots, and exhibit moderate seasonality.

\section{Implementation Details}
\label{app:impl}

\paragraph{Tabular baselines.}
We trained \textbf{FT-Transformer}, \textbf{CatBoost}, and \textbf{TabM} on the same preprocessed dataset (categorical features: \texttt{season}, \texttt{time}; numerical meteorological predictors; target: multi-level wind profiles). The dataset was split 80/20 with stratification by \texttt{season $\times$ time}; targets were standardized and inverse-scaled for evaluation. Total compute budget across all experiments (per-country and global runs combined) is \textbf{653 GPU-hours}.

\textbf{FT-Transformer:} $d=256$, 6 encoder layers, 16 attention heads, feed-forward size 512, dropout 0.2, GELU activation, [CLS]-token aggregation; \texttt{AdamW} with $lr=3{\cdot}10^{-4}$, weight decay $10^{-4}$, batch size $30\,000$, early stopping with patience 250. Model size $\approx 3$--$4$M parameters. Training time: 5\,h (global), $\approx$1\,h per country.
\textbf{CatBoost:} 3000 iterations, depth 8, $lr=0.03$, objective \texttt{MultiRMSE}, early stopping with \texttt{od\_wait=200}, GPU backend; up to 256 leaves per tree. Training time: 2.3\,h (global), $\approx$1\,h per country.
\textbf{TabM:} 3 blocks of size 512, ensemble size $k=32$, dropout 0.1, numeric embeddings with piecewise-linear bins (64) and $d_{\mathrm{emb}}=16$; \texttt{AdamW} with $lr=2{\cdot}10^{-3}$, weight decay $3{\cdot}10^{-4}$, batch size $30\,000$, early stopping with patience 250. Training time: 4\,h (global), $\approx$1\,h per country.

\paragraph{FNO post-hoc corrector.}
The corrector summarized in Section~\ref{sec:results} (Figure~\ref{fig:fno}) is applied identically to all three frozen tabular baselines; only the TabM curve is shown in the main body for clarity. Given scaled targets $Y_{\text{train}}^{(t)}, Y_{\text{test}}^{(t)}$ and frozen baseline predictions $\hat{Y}$, we (1) tile $\hat{Y}$ along a synthetic 1D axis of fixed length $S=64$, producing tensors $X \in \mathbb{R}^{B\times C\times S}$ with $C = N_{\text{targets}}$; (2) tile the ground-truth targets to match; (3) train an FNO to minimize MSE on the tiled tensors; (4) average over $S$ at inference to recover predictions of shape $[B, C]$. The FNO configuration is held constant across baselines: $n_{\text{modes}}\!=\![16]$, hidden channels $64$, lifting/projection channels $256$, $n_{\text{layers}}\!=\!4$, GELU nonlinearity, linear skip connections, no normalization, factorized spectral implementation (\texttt{factorization=tucker}, \texttt{rank}=0.5, \texttt{decomposition=cp}). Optimization uses \texttt{AdamW} ($lr=10^{-3}$, weight decay $10^{-4}$), batch size $10$k, up to 1000 epochs with early stopping (patience 100), and \texttt{ReduceLROnPlateau} (factor 0.5, patience 20, cooldown 10, $lr_{\min}\!=\!10^{-6}$). We keep the 80/20 split, apply z-score scaling to $Y$, and inverse-transform predictions for reporting.

\section{Reproducibility Statement}

\textbf{Data:} IGRA radiosonde measurements are public (NOAA NCEI, \url{https://www.ncei.noaa.gov/data/ecmwf-global-upper-air-bufr/archive/}); preprocessing, $\HABL$ definition, and spline interpolation are described in Section~\ref{sec:dataset}. \textbf{Models:} hyperparameters and architectures are listed in Appendix~\ref{app:impl}; the FNO post-hoc corrector is fully specified there. \textbf{Variance:} RMSE values in Section~\ref{sec:results} are averaged across 5 random seeds; standard deviations appear in Table~\ref{tab:rmse_summary}. \textbf{Code:} an anonymized repository with preprocessing scripts, model implementations, and configurations will be released to reviewers and made public upon publication.

\end{document}